\documentclass[twocolumn]{svjour3}
\RequirePackage{fix-cm}
\newcommand{\makeheadbox}{}
\usepackage{graphicx}
\usepackage{amsmath,amssymb,amsfonts}
\usepackage{microtype}
\usepackage{mathrsfs}
\usepackage[table]{xcolor}
\usepackage{textcomp}
\usepackage{manyfoot}
\usepackage{adjustbox}
\usepackage{booktabs}
\usepackage{tabularx}
\usepackage{multirow}
\usepackage{array}
\usepackage{makecell}
\usepackage{float}
\usepackage{wrapfig}
\usepackage{placeins}
\usepackage{rotating}
\usepackage{subcaption}
\usepackage{algorithm}
\usepackage{algorithmicx}
\usepackage{etoolbox}
\usepackage{algpseudocode}
\usepackage{listings}
\usepackage{dirtytalk}
\usepackage{soul}
\usepackage{framed}
\usepackage{comment}
\usepackage{titlesec}
\usepackage[numbers]{natbib}
\usepackage{url}
\usepackage[
    colorlinks=true,
    linkcolor=black,
    citecolor=black,
    urlcolor=blue
]{hyperref}

\IfFileExists{bbm.sty}
  {\usepackage{bbm}}
  {}

\begin{document}

\title{Enhancing Multi-Region Stylization with
Interior-Guided Boundary Repair}

\titlerunning{Interior-Guided Boundary Repair}
\authorrunning{Hong-Son Nguyen and Thi-Ngoc-Hanh Le}

\author{Hong-Son Nguyen \and Thi-Ngoc-Hanh Le}

\institute{%
Hong-Son Nguyen \at
International University, VNU-HCM\\
\email{sonhongnguyen.researcher@gmail.com}
\and
Thi-Ngoc-Hanh Le (corresponding author) \at
International University, VNU-HCM\\
\email{ltnhanh@hcmiu.edu.vn}
}
\date{}
\maketitle
\begin{abstract}Region-based neural style transfer enables fine-grained artistic control by allowing independent stylization of semantic image regions. However, compositing these regions often leads to boundary artifacts, degrading visual quality. We propose Interior-Guided Boundary Repair (IGBR), a lightweight and model-agnostic method that improves boundary handling in multi-region stylization. IGBR repairs boundary pixels using interior-guided propagation and applies inward, distance-based blending restricted to object-background boundaries, preventing inter-object style leakage. The method is derived from a region-wise constrained formulation with a closed-form solution and can be seamlessly integrated into existing stylization pipelines without retraining. To evaluate efficiency of our IGBR, we introduce quantitative metrics that measure boundary consistency, gradient artifacts, inter-object leakage, and interior preservation without requiring annotated stylized images. Our experiments and evaluations demonstrate that the proposed IGBR consistently produces plausible boundaries, outperforming prior blending techniques in boundary consistency, gradient stability, and interior preservation. The code is available at
\url{https://github.com/Son-SDT/IGBR}.
\end{abstract}

\section{Introduction}
\begin{figure}[t]
    \centering     
    \includegraphics[width=\linewidth]{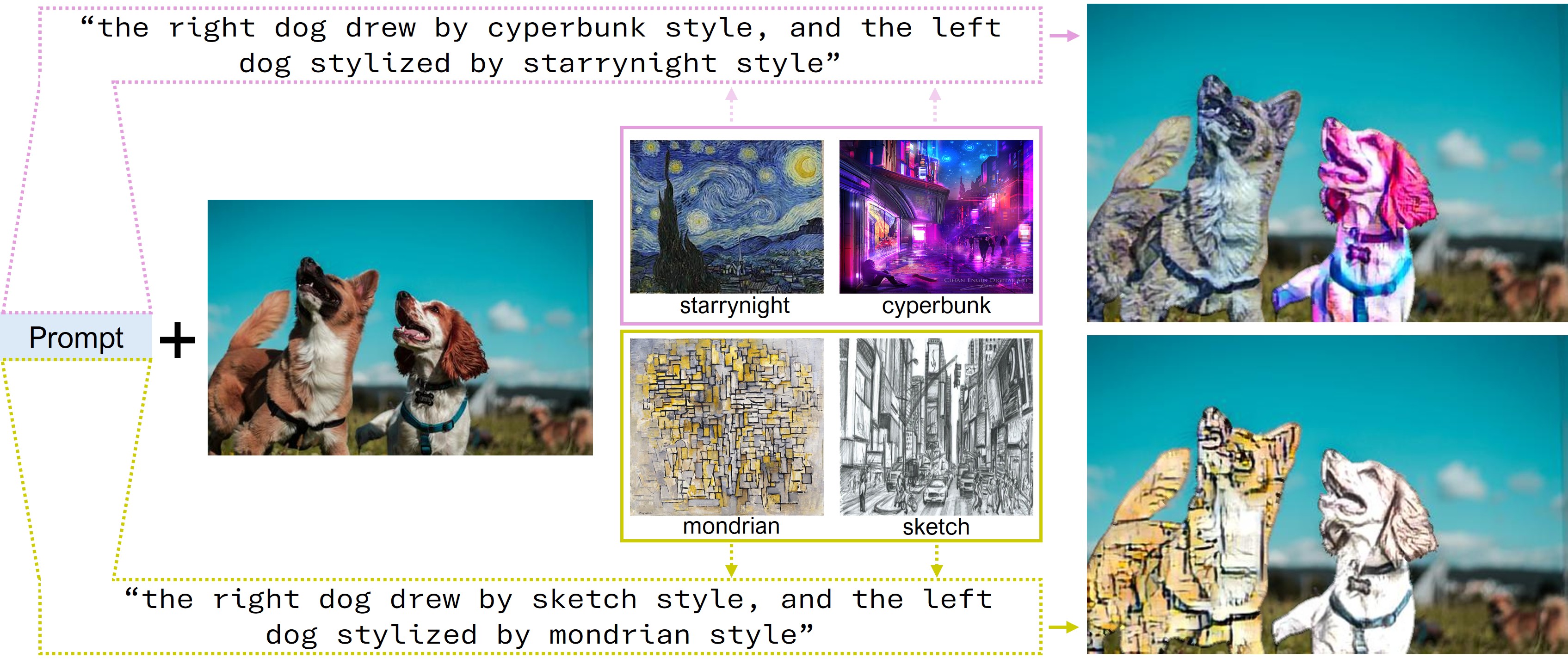}
    \caption{\small Our framework enables independent stylization of multiple semantic regions with different styles while preserving region boundaries and reducing cross-region interference.}
    \label{teaser}
\end{figure}

Neural style transfer (NST) has become a fundamental tool for artistic image synthesis, allowing the transformation of photographs into stylized artworks by transferring visual characteristics from reference styles \cite{gatys2016image,johnson2016perceptual,huang2017arbitrary, xu2024image, wu2024lighting}. While early methods focus on global image stylization, recent applications increasingly demand region-aware and multi-region stylization, where different objects or semantic regions within an image are stylized independently according to user intent  \cite{champandard2016semantic,luan2017deep,kolkin2019style}. These capabilities are important for several applications, such as interactive content creation, digital art, and personalized media generation \cite{han20153d, wang2020photographic, ou2025wdanet}. Fig.\ref{teaser} exhibits such capability, which is generated by our proposed framework.

Despite progress in region-based stylization, boundary handling remains challenging. Independently stylized regions often produce halos, abrupt transitions, texture bleeding, and inter-object style leakage after composition, especially in images with fine structures or multiple stylized objects. Existing methods usually rely on global blending strategies, such as alpha blending, Poisson blending, or matting-based techniques \cite{perez2023poisson,levin2008closed}. However, these methods treat all boundaries uniformly and do not distinguish object-background boundaries from interfaces between different stylized regions. Consequently, style information may propagate across semantically distinct objects, leading to inconsistent or visually implausible results \cite{luan2017deep,selim2016painting}.

In this work, we found that boundary handling in region-based stylization should not be treated as a generic post-processing step, but rather as a region-wise constrained problem that explicitly accounts for semantic structure. We observe that, perceptually, boundary pixels should primarily reflect the appearance of their own object interior, while stylization strength should increase progressively toward the interior and remain unaffected in deep interior regions. This observation motivates us formulating the boundary correction in blending as directional, localized, and interior-guided approach, rather than global and symmetric.

Based on this insight, we propose an \emph{Interior-Guided Boundary Repair} (\textbf{IGBR}) algorithm, a lightweight yet principled algorithm for boundary handling in multi-region stylization. IGBR operates by first repairing boundary pixels using nearest interior propagation to enforce interior consistent boundary conditions, and then applying an inward, distance-controlled feathering that smoothly transitions from the original image to the stylized composite. More specifically, blending is applied exclusively at object-background boundaries, while boundaries between stylized objects are explicitly excluded, preserving semantic isolation between regions.

Unlike prior work, IGBR is model-agnostic and does not require retraining or modification of existing neural style transfer models \cite{johnson2016perceptual,huang2017arbitrary}. It can be seamlessly integrated with any region-based stylization pipeline and admits a closed-form solution derived from a constrained optimization formulation. This design yields strong
theoretical guaranties: boundary effects are localized, inter-object style leakage is suppressed, and interior stylization is provably preserved. The extensive qualitative and quantitative evaluations presented in our work demonstrate that IGBR consistently outperforms classical blending methods in boundary consistency, gradient stability, and interior preservation, while avoiding the common artifacts observed in existing approaches. Together, these results show that explicitly formulating boundary handling as a region-wise constrained problem offers a robust and effective solution for multi-region stylization.

\noindent Our main contributions to this work are given follows.
\begin{itemize}
    \item [-] A novel formulation of boundary handling in region-based stylization as a region-wise constrained optimization problem.
    \item [-] The proposed Interior-Guided Boundary Repair (IGBR) algorithm, which enforces interior-consistent boundaries and directional blending without global smoothing.
    \item [-] A set of task-specific quantitative metrics that directly evaluate boundary consistency, inter-object style leakage, gradient artifacts, and interior preservation without requiring ground-truth stylized images.
    \item [-] Extensive experiments demonstrating improved boundary quality and semantic coherence across diverse multi-region stylization scenarios.
\end{itemize}

\section{Related Work}
\textbf{Neural Style Transfer.}
Neural style transfer (NST) was introduced by Gatys et al.~\cite{gatys2016image} as the optimization of CNN-based content and style statistics. Later works improved efficiency and controllability using feed-forward networks and perceptual losses~\cite{johnson2016perceptual,ulyanov2016texture,11365097, zhang2023portrait,le2023structure,sun2025image}, and enabled arbitrary style transfer through adaptive normalization~\cite{huang2017arbitrary,li2017demystifying,park2019arbitrary,zhang2024s2wat,yu2024multi,dou2025mla,ge2024arbitrary}. Other studies enhance style representation with optimal transport~\cite{kolkin2019style}, attention mechanisms~\cite{svoboda2020stylizing}, and saliency- or SIFT-guided fusion~\cite{sun2025image}. Although effective for global stylization, these methods mainly apply style at the image level and do not explicitly address boundary artifacts caused by independent region-wise stylization.

\textbf{Region-Based and Semantic Stylization. }To provide localized control, several methods incorporate semantic or instance-level cues into NST. Champandard~\cite{champandard2016semantic} and Selim et al.~\cite{selim2016painting} introduced semantic style transfer by constraining style losses within segmented regions. Luan et al.~\cite{luan2017deep} used semantic segmentation and photorealism constraints to preserve image structure in photo style transfer. Later works further explored soft masks and layer-wise constraints for region-based stylization~\cite{gatys2017controlling,mechrez2017contextual}. However, these methods mainly control \say{where} styles are applied, while paying less attention to \say{how} independently stylized regions are recomposed. As a result, boundary handling is often delegated to generic blending operations, which can produce visible artifacts when multiple stylized regions coexist. 

\textbf{Image Compositing and Blending.} Classical compositing methods are commonly used for boundary handling. Alpha blending is efficient but prone to halos, while Poisson editing \cite{perez2023poisson} improves smoothness but may introduce texture leakage. Matting methods, such as closed-form matting and its extensions \cite{levin2008closed, li2023deep}, enhance boundary fidelity but often require separable foreground-background appearances or accurate trimaps. Edge-aware filters~\cite{he2010guided} preserves local edges but lack semantic constraints. Although widely adopted in stylization and photo editing~\cite{luan2017deep,selim2016painting}, these methods treat boundaries uniformly, leading to over-smoothing, texture bleeding, or unintended style mixing in multi-region stylization~\cite{tanaka2019style,liu2021towards}.

\textbf{Boundary-Aware Stylization and Editing.} Several studies have emphasized boundary preservation in image editing and stylization~\cite{he2010guided,cheng2015deep,tanaka2019style,kotovenko2019content,liu2021towards,junjie2023image,hu2024msembgan,pan2025clpfusion}. Edge-aware and guided filtering preserve strong edges during smoothing~\cite{tomasi1998bilateral,he2010guided}, while gradient-domain constraints improve structural consistency~\cite{tanaka2019style}. Cheng et al.~\cite{cheng2015deep} reformulated colorization with deep learning and joint bilateral filtering to reduce artifacts. Recent works further explore controllable or region-aware stylization through spatial constraints~\cite{kotovenko2019content,liu2021towards}, defocused-image boundary ringing removal~\cite{junjie2023image}, diffusion-based landscape style transfer~\cite{pan2025clpfusion}, and region-aware texture synthesis with adversarial learning~\cite{hu2024msembgan}. However, these methods generally do not explicitly guarantee boundary behavior or distinguish object-background boundaries from inter-object interfaces. Unlike these, our method formulates multi-region recombination as a region-wise constrained refinement problem, making it model-agnostic and free from additional training or annotated data.

In contrast to above methods, we formulate boundary handling in region-based stylization as a region-wise constrained problem. The proposed Interior-Guided Boundary Repair (IGBR) combines interior-consistent repair with inward feathering restricted to object-background boundaries. This closed-form, model-agnostic formulation preserves semantic isolation between stylized regions and provides a principled solution with explicit quantitative analysis for multi-region style transfer.

\section{Methodology}
\subsection{Problem Definition}
We investigate multi-region neural style transfer, where an input image is decomposed into semantically distinct regions and each region is assigned an individual artistic reference. Given an input image $\mathcal{I}$, semantic masks $\{M_i\}_{i=1}^{N}$, and style references $\{S_i\}_{i=1}^{N}$, our goal is to generate a single stylized image that preserves the semantic structure of $\mathcal{I}$ while reflecting the style specified for each region.

Unlike global style transfer, multi-region stylization must handle independently stylized regions within the same image. When these regions are recomposed, boundary artifacts often appear, including color discontinuities, halos, texture misalignment, and inconsistent brush patterns. These defects reduce perceptual quality even when the interior of each region is successfully stylized. Existing composition methods, such as alpha blending, gradient-domain blending, and matting-based refinement, are not designed for heterogeneous region-wise styles. They usually assume compatible appearance statistics across neighboring regions and allow information exchange across boundaries. In multi-style stylization, such propagation may cause style leakage and semantic ambiguity.

\begin{figure}[t]
    \centering     
    \includegraphics[width=0.99\linewidth]{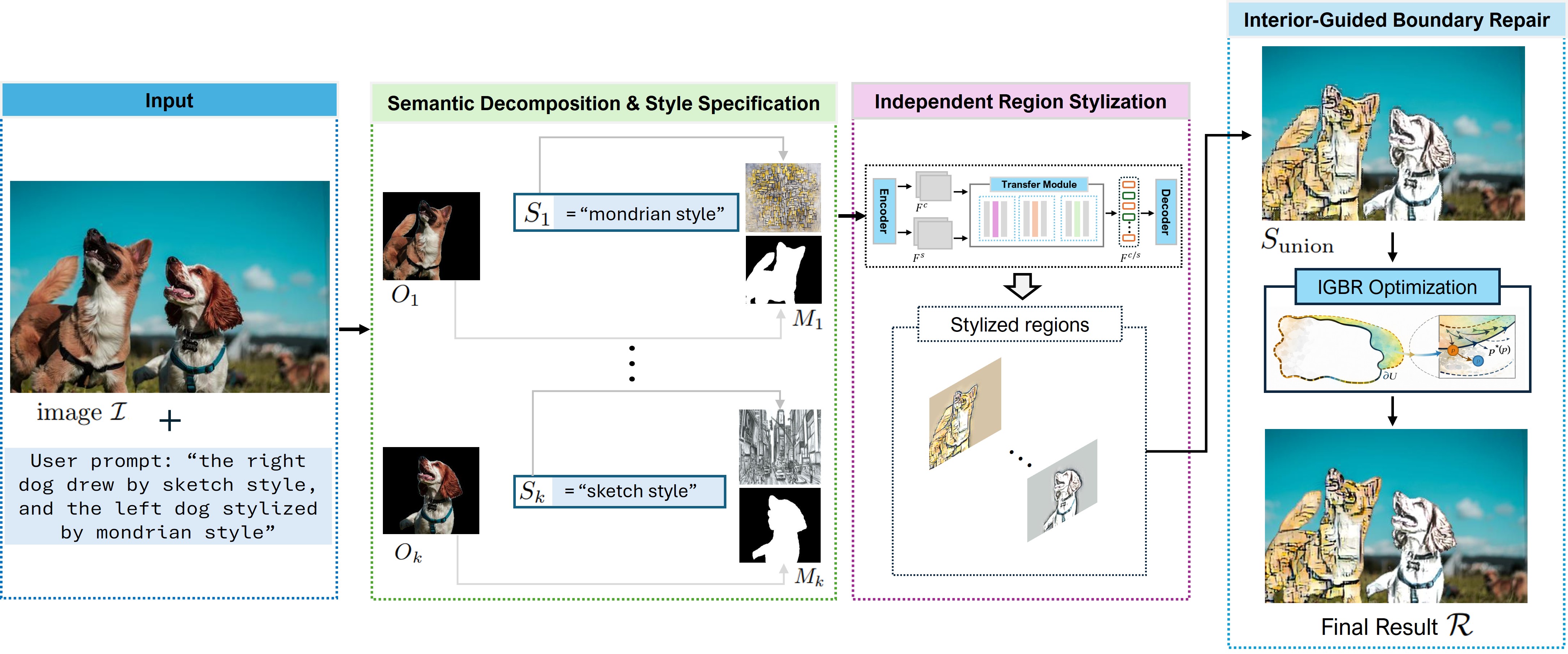}
    \caption{\small Overview of the proposed multi-region stylization framework}
    \label{framework}
\end{figure}

To address this issue, we formulate boundary correction as a region-wise constrained refinement problem rather than a global blending task. As shown in Fig.\ref{framework}, our framework contains three stages: semantic decomposition and style specification, independent region stylization, and Interior-Guided Boundary Repair (IGBR). The first two stages provide flexible region-wise control, while IGBR is the core contribution that repairs boundary artifacts under semantic isolation constraints.

Instead of blending information across adjacent regions, IGBR regularizes the boundary from the interior side of each stylized union. Reliable interior appearance is propagated toward the boundary, and the correction is restricted to the object-background interface. This inward-only design suppresses boundary artifacts without contaminating neighboring regions or weakening the stylized interior. As a result, boundary handling becomes a principled refinement stage that is compatible with different stylization backbones, masks, and region configurations.

\subsection{Semantic Decomposition and Style Specification}
This stage converts high-level user intent into explicit region-wise stylization constraints. Given an input image $\mathcal{I}$ and a user prompt, our goal is to identify which semantic regions should be stylized, which artistic style should be assigned to each region, and where these regions are located in the image domain. This stage provides the semantic partition required for independent stylization and for the constrained Boundary Repair introduced in the later stage.

To parse the user instruction, we employ a large language model~\cite{team2023gemini} to extract semantic objects and their corresponding style descriptions. The prompt is therefore transformed into a finite set of object-style pairs: $\{(O_k, S_k)\}_{k=1}^{N}$,where $O_k$ denotes the $k$-th semantic object and $S_k$ denotes its associated style descriptor. Each pair $(O_k, S_k)$ defines an independent stylization constraint. This representation separates semantic intent from low-level image processing, allowing the framework to support different object categories, user instructions, and artistic styles in a modular manner.

For each semantic object $O_k$, we estimate a binary region mask $M_k$ using a segmentation model~\cite{kirillov2023segment}. The mask specifies the spatial support of the object in the input image:
\begin{equation}
    M_k(p) =
    \begin{cases}
        1, & \text{if pixel } p \text{ belongs to object } O_k, \\
        0, & \text{otherwise}.
    \end{cases}
\end{equation}
The set of masks $\{M_k\}_{k=1}^{N}$ decomposes the image into semantic regions that can be stylized independently. This pixel-level decomposition is important because the following stages require explicit spatial constraints to preserve semantic isolation and prevent style propagation across neighboring regions.

For style specification, each descriptor $S_k$ is associated with a concrete style reference image $I_k^s$. In practice, $S_k$ is used to query a style repository, and the retrieved image serves as the style exemplar for region $k$. The resulting triplet $(O_k, M_k, I_k^s)$ defines what should be stylized, where the stylization should be applied, and which visual appearance should be transferred.

Overall, semantic decomposition and style specification provide the structural and stylistic constraints of the proposed multi-region stylization framework. By separating prompt interpretation, mask estimation, and style reference retrieval from the stylization process itself, this stage improves the modularity of the framework and allows multiple styles to coexist within a single image. The obtained masks and style references are then used for independent region stylization and for the Interior-Guided Boundary Repair described in Section~3.4.

\subsection{Independent Region Stylization}
This stage generates region-specific stylized outputs based on the semantic decomposition and style specification obtained in the previous stage. Given the set of semantic masks $\{M_k\}_{k=1}^{N}$ and the corresponding style reference images $\{I_k^s\}_{k=1}^{N}$, each region is stylized independently to produce a region-wise stylized result $\hat{I}_k$. The purpose of this stage is to ensure that each semantic region can fully express its assigned artistic style without being affected by neighboring regions.

Jointly stylizing all regions in a single optimization or generation process may cause incompatible styles to interact in the feature space. This interaction can weaken the stylistic identity of individual regions, especially when adjacent regions are assigned visually contrasting styles. Therefore, we separate region stylization from region composition. This design allows multiple artistic styles to coexist in the same image while preserving the local semantic meaning of each region.

In our implementation, we employ S2WAT~\cite{zhang2024s2wat} as the neural style transfer backbone to perform region-conditioned stylization. For each region $k$, the stylization process is defined as:
\begin{equation}
    \hat{I}_k = \mu(\mathcal{I}, I_k^s, M_k),
\end{equation}
where $\mu(\cdot)$ denotes a neural style transfer model, $\mathcal{I}$ is the input image, $I_k^s$ is the style reference image assigned to region $k$, and $M_k$ is the corresponding semantic mask. Using an existing stylization backbone allows the proposed framework to focus on the boundary inconsistency problem rather than on redesigning the stylization model itself. It is also worth noting that the formulation is not restricted to S2WAT. Any region-aware style transfer model that produces per-region stylized outputs can be integrated into the proposed framework without changing the subsequent Boundary Repair stage.

After obtaining all region-wise stylized outputs, we construct a preliminary composite image by placing each stylized region back into its semantic support:
\begin{equation}
    S_{\text{union}} =
    \sum_{k=1}^{K} \hat{I}_k \odot M_k
    +
    \mathcal{I} \odot
    \left(1 - \bigcup_{k=1}^{K} M_k\right),
\end{equation}
where $\odot$ denotes element-wise multiplication. The first term combines all stylized regions, while the second term preserves the original appearance of pixels that do not belong to any stylized region. This operation produces a complete stylized image while maintaining the semantic assignment defined by the user prompt.

However, independent region stylization does not explicitly model boundary compatibility in the preliminary composite $S_{\text{union}}$. Therefore, visible artifacts such as color discontinuities, halos, texture breaks, and inconsistent brush patterns may appear near region interfaces, especially when adjacent regions use contrasting style references. Rather than constraining the stylization backbone to resolve these artifacts, we isolate them after composition and address them through a separate constrained refinement stage. This design keeps the framework modular, preserves the interior stylization of each region, and delegates boundary correction to the proposed Interior-Guided Boundary Repair described in Section~3.4.

\subsection{Interior-Guided Boundary Repair (IGBR)}
This stage is the core component of the proposed framework. It aims to correct boundary artifacts introduced by independent region stylization while preserving the semantic separation among different objects. Instead of treating boundary correction as a conventional blending problem, we formulate it as a region-wise constrained refinement process defined on the union of all stylized regions. The formulation is guided by two principles. First, semantic isolation prevents information from propagating across object boundaries. Second, inward-only regularization restricts the correction direction from the outer object boundary toward the interior, thereby suppressing halo artifacts without weakening the stylized content inside each region.

Given the image $\mathcal{I}$ and let $\{M_k\}_{k=1}^{N}$ be the set of semantic masks. We define the union of all stylized regions as:
\begin{equation}
    U = \bigcup_{k=1}^{N}\{p \in \mathcal{I} \mid M_k(p)=1\}.
\end{equation}
The outer boundary of the union region is then defined as:
\begin{equation}
    \Phi =  \{p \in U \mid \exists q \in \mathcal{N}(p),\ q \notin U\},
\end{equation}
where $\mathcal{N}(p)$ denotes the local pixel neighborhood of $p$. This definition is important because it only considers the interface between the stylized foreground union and the non-stylized background. Boundaries between semantic objects inside $U$ are not treated as blending boundaries. As a result, the method avoids cross-object propagation and preserves the visual identity of independently stylized regions.
\begin{figure}[H]
    \centering 
    \includegraphics[width=\linewidth]{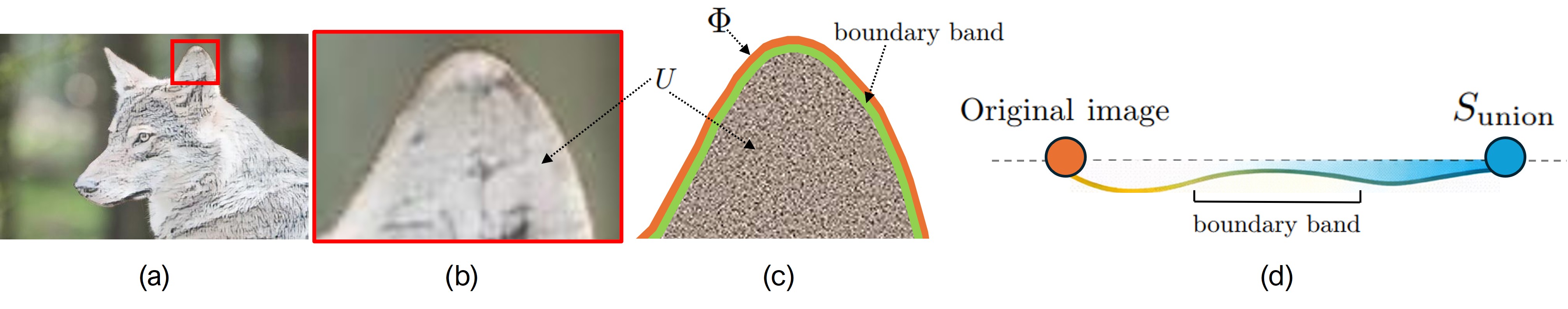}
    \caption{\small Visualization the concept in our IGBR }
    \label{IGBR_visualize}
\end{figure}
Let $S_{\text{union}}$ denote the preliminary composite image obtained from the independent region stylization stage. Although $S_{\text{union}}$ contains the desired region-wise styles, pixels near $\Phi$ may contain halo effects, color discontinuities, or unstable textures caused by mask edges and independent stylization. To localize the correction, we first compute an inward distance transform from the union boundary:
\begin{equation}
    D_U(p) = \operatorname{dist}(p,\Phi),
    \quad p \in U.
\end{equation}
Given a repair width $\tau$, we define a narrow inward repair band:
\begin{equation}
    \mathcal{B}_{\tau}
    =
    \{p\in U \mid D_U(p)<\tau\}.
\end{equation}
The repair operation is applied only to pixels in $\mathcal{B}_{\tau}$, while pixels outside this band are kept unchanged before feathering. For each $p\in\mathcal{B}_{\tau}$, we identify its nearest reliable interior pixel:
\begin{equation}
    p^{*}(p)
    =
    \arg\min_{q \in U,\ D_U(q)\geq \tau}
    \|p-q\|_2,
\end{equation}
and update the preliminary composite by:
\begin{equation}
    S_{\text{union}}(p)\leftarrow S_{\text{union}}(p^{*}(p)).
\end{equation}

This operation replaces unstable boundary appearance using information propagated from the interior of the same stylized union. Since $p^{*}(p)$ is constrained to lie inside $U$, the repair process does not borrow colors or textures from the background. Moreover, because the correction is defined with respect to the union boundary $\Phi$, it does not smooth across internal object interfaces. This yields an interior-consistent boundary condition before the final inward feathering step.

Using the same inward distance transform, we define the inward blending weight as:
\begin{equation}
    \alpha(p) =\min\left(\frac{D_U(p)}{\tau},1\right).
\end{equation}
The weight $\alpha(p)$ increases monotonically with the distance from the boundary. When $p$ is close to $\Phi$, $D_U(p)\simeq 0$ and $\alpha(p)\simeq 0$, so the output remains closer to the original image. When $p$ is sufficiently deep inside the stylized region, $D_U(p)\geq \tau$ and $\alpha(p)=1$, so the output fully preserves the stylized composite. Therefore, the correction is localized to a narrow inward band, while the interior stylization remains unchanged.

Fig.~\ref{IGBR_visualize} illustrates the main behavior of the proposed IGBR algorithm. Given a stylized object, boundary artifacts can be observed around the object contour, such as the highlighted region near the fox ear. The union region $U$ and its outer boundary $\Phi$ define where the method is allowed to operate. The boundary-adjacent pixels are first repaired using interior-guided propagation and are then blended according to their inward distance from $\Phi$. As a result, pixels close to the boundary are stabilized by the original image, whereas pixels deeper inside the object converge to the stylized composite. This process suppresses halo artifacts and abrupt boundary transitions without altering the semantic layout or weakening region-specific stylization.

After the interior-guided repair, the final output image $\mathcal{R}$ is obtained by solving the following constrained optimization problem:
\begin{equation}
\begin{aligned}
\min_{\mathcal{R}:\Omega\rightarrow\mathbb{R}^3}\quad
& \sum_{p\in U}\Big(
\alpha(p)\,
\|\mathcal{R}(p)-S_{\text{union}}(p)\|_2^2 \\
& \qquad\qquad
+ (1-\alpha(p))\,
\|\mathcal{R}(p)-\mathcal{I}(p)\|_2^2
\Big)\\ \text{ s.t. }
& \mathcal{R}(p)=\mathcal{I}(p),\text{ } \forall p\notin U.
\end{aligned}
\end{equation}
This objective balances two terms. The first term preserves the repaired stylized composite, while the second term anchors the output to the original image near the union boundary. The spatially varying weight $\alpha(p)$ controls this balance according to the inward distance. Since the objective is separable across pixels, it admits the following closed-form solution:
\begin{equation}
\mathcal{R}(p)=
\begin{cases}
\alpha(p)\,S_{\text{union}}(p) +
(1-\alpha(p))\,\mathcal{I}(p),
& p\in U,\\\mathcal{I}(p),& p\notin U.
\end{cases}
\end{equation}
The solution produces a smooth and spatially controlled transition from the original image to the stylized composite as the inward distance increases. Pixels adjacent to the boundary remain visually stable and close to the source image, while pixels inside the union region retain the full stylization effect. Consequently, boundary handling is localized, directional, and semantically constrained. The pseudocode of the proposed IGBR strategy is summarized in Algorithm~\ref{IGBR_alg}.

\begin{algorithm}[t]
\caption{\small Interior-Guided Border Repair}
\label{IGBR_alg}
\begin{algorithmic}[1]
\Statex \textbf{Input:} Input image $\mathcal{I}$; stylized regions $\{\hat{\mathcal{I}}_k\}_{k=1}^{N}$; masks $\{M_k\}_{k=1}^{N}$; feather width $\tau$
\Statex \textbf{Output:} Final stylized image $\mathcal{R}$

\State Construct the union mask $U$ by Eq.(4)
\State Construct the preliminary stylized composite $S_{\text{union}}$ by Eq.(3)
\State Extract the outer union boundary $\Phi$ by Eq.(5)
\State Compute the inward distance transform $D_U(p)$ Eq.(6)
\State Define the narrow inward repair band:
$\mathcal{B}_{\tau}$ by Eq.(7)
\For{each pixel $p \in \mathcal{B}_{\tau}$}
    \State Find the nearest reliable interior pixel
    $p^{*}(p)$ Eq.(8)
    \State Repair the preliminary composite \\ 
    $\qquad S_{\text{union}}(p) \leftarrow
        S_{\text{union}}(p^{*}(p))$
\EndFor
\State Compute the inward feathering weight
    $\alpha(p)$ by Eq.(10)
\For{each pixel $p\in\mathcal{I}$}
    \If{$p\in U$}
        \State \[\mathcal{R}(p) =
        \alpha(p)S_{\text{union}}(p)
        +(1-\alpha(p))\mathcal{I}(p) \]
    \Else
        \State $\mathcal{R}(p)\leftarrow \mathcal{I}(p)$
    \EndIf
\EndFor
\State \Return $\mathcal{R}$
\end{algorithmic}
\end{algorithm}

\section{Experiments}
\subsection{Implementation Details}
All experiments were conducted on a desktop PC equipped with an Intel Core i7 CPU, 16~GB RAM, and an NVIDIA GeForce RTX 3070 GPU. The overall framework is implemented in Python. Object masks are obtained using a pre-trained segmentation model and are used to define region supports for independent stylization. The proposed Interior-Guided Boundary Repair (IGBR) is implemented as a lightweight post-processing module operating directly in image space and does not require additional network training.

The inward distance transform and boundary-band computation are implemented using standard Euclidean distance transforms, while nearest interior pixel queries are resolved via distance-based lookup. Unless otherwise specified, the feathering width parameter is set to $\tau=7$ pixels for images of resolution $512\times512$, and scaled proportionally for other resolutions. All experiments are performed on normalized RGB images. The demo video of our framework can be explored here \footnote{https://youtu.be/Fn4K0Dsr0I8}.

\subsection{Performance Analysis}
\subsubsection{Qualitative evaluations}
\begin{figure}[t]
    \centering 
    \includegraphics[width=0.9\linewidth]{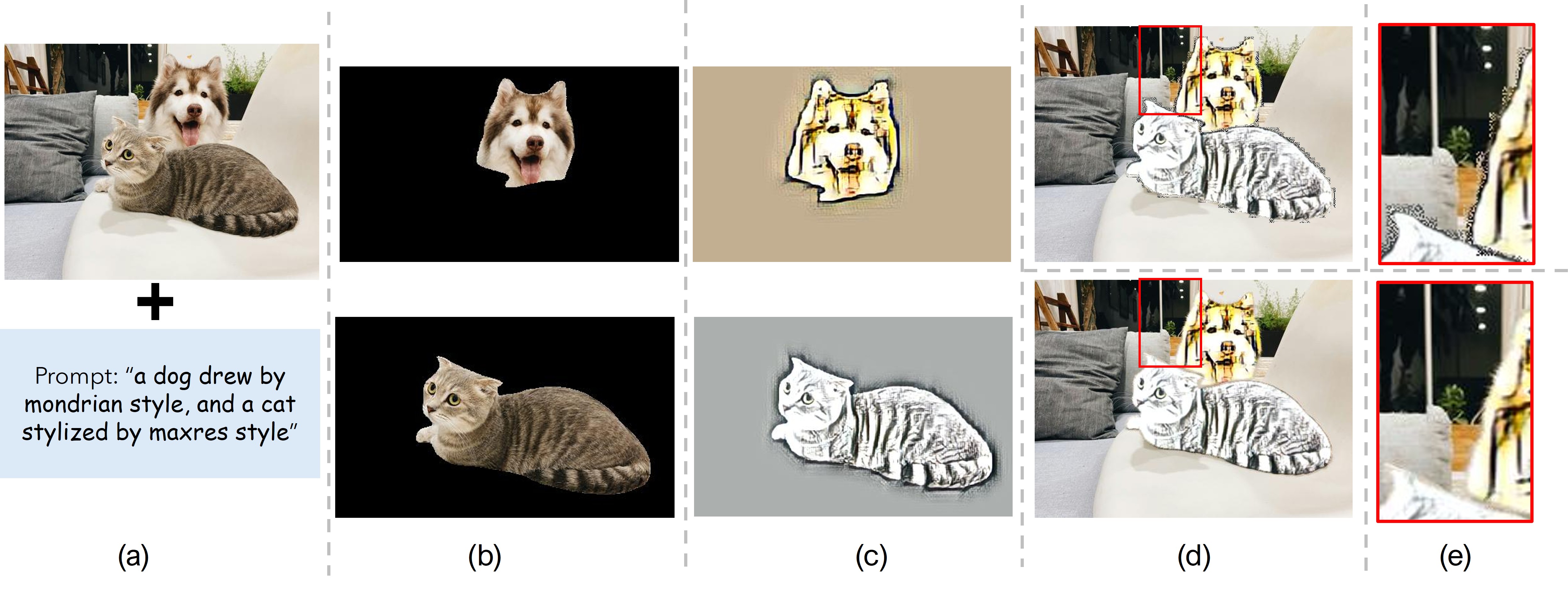}
    \caption{\small (a) inputs, (b) segmented objects, (c) stylized objects, (d) results by Alpha blending (upper) and our IGBR (lower), (e) zoom-in of regions in (d).}
    \label{results}
\end{figure}

Figure \ref{results} illustrates a representative multi-region stylization result and highlights the effectiveness of the proposed IGBR. Fig.\ref{results}(a) shows the input image and prompt specifying heterogeneous styles for two semantic objects. This setting is challenging due to the strong stylistic contrast between the sketch-style dog and the cubist-style cat. Fig.\ref{results}(b) presents the segmented object regions, which enable independent stylization while preserving spatial alignment with the original scene. Fig.\ref{results}(c) are the independently stylized objects produced by the neural style transfer model. While each object accurately reflects its target style, these intermediate results do not encode boundary consistency, as stylization is performed in isolation.

The performance of our IGBR algorithm is evident in Fig.\ref{results}(d)-(e), which compares pure blending (upper row) with ours strategy (lower one). We can observe the effect the highlighted region (red and green rectangles). Obviously, pure blending produces visible boundary artifacts. These reduce semantic separation and visual quality. In contrast, our IGBR can maintain smooth, plausible, and perceptual consistent transitions at the object boundaries. 

\begin{figure}[t]
    \centering 
    \includegraphics[width=\linewidth]{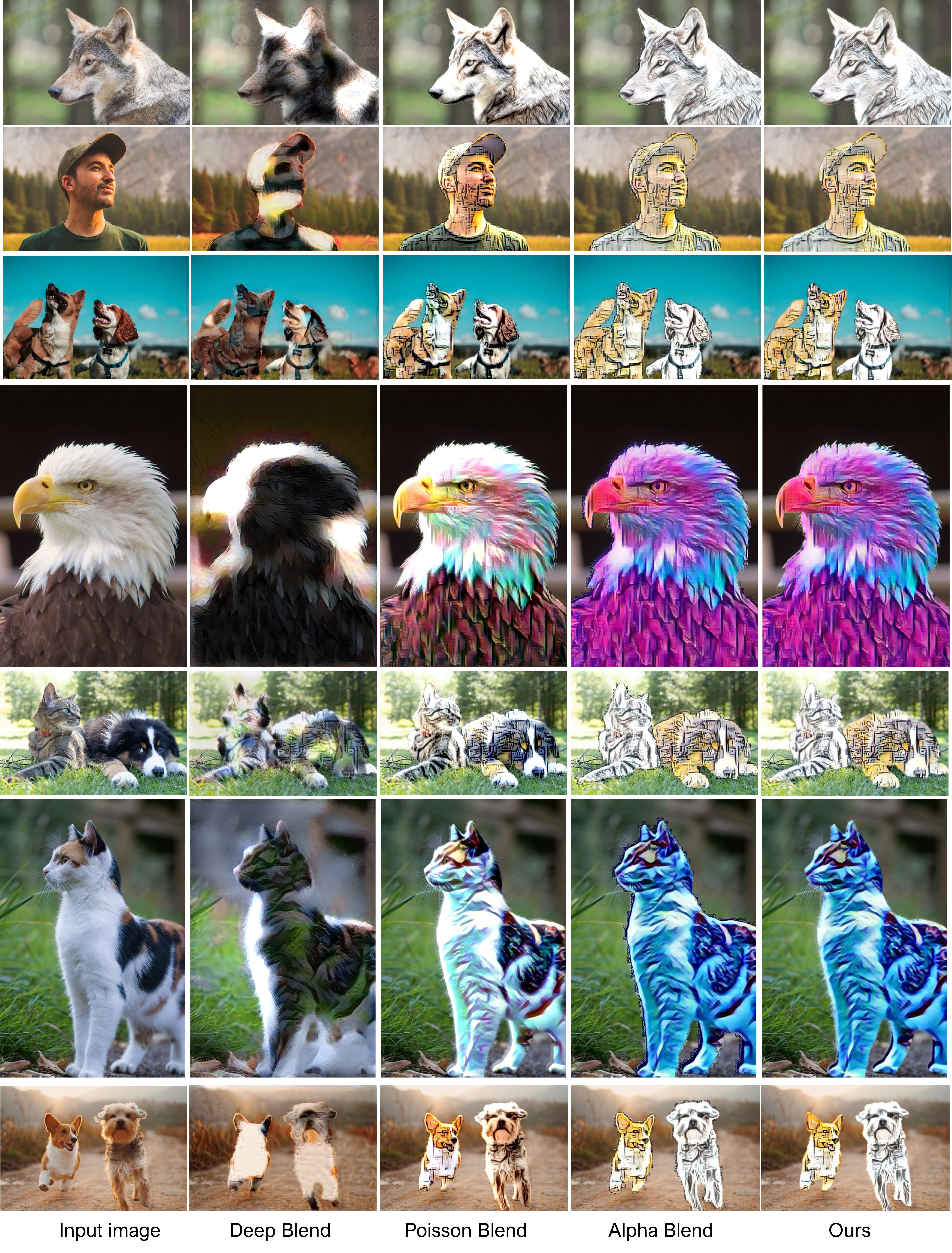}
    \caption{\small Visual comparisons between our IGBR and the prior works.}
    \label{comparisons}
\end{figure}

To demonstrate more, we exhibit comparisons on several samples between our IGBR and Deep Blending \cite{zhang2020deep}, Poisson Blending \cite{perez2023poisson}, and Alpha Blending in Fig. \ref{comparisons}. These cases are particularly challenging due to the strong contrast between stylized foreground objects and natural backgrounds. We can see that, Deep Blend produces visually smooth results but often exhibits over-smoothing and style leakage near object boundaries, especially along high-frequency structures such as fur. Poisson blending improves gradient continuity but introduces unintended texture propagation between foreground and background, resulting in inconsistent object appearance. Alpha blending preserves object isolation but suffers from abrupt and visually distracting boundary discontinuities due to the lack of boundary regularization. In contrast, the results generated by our IGBR consistently preserves semantic separation while maintaining smooth and perceptually coherent object–background transitions. Stylized textures remain confined to object interiors and no cross-region contamination is observed. This behavior directly follows from the constrained optimization in Eq.(11)-(12), where boundary pixels are repaired using interior-guided propagation and smoothing is applied only inward via distance-based weighting. Overall, this comparison demonstrates that explicitly modeling boundary handling as a region-wise constrained problem enables more robust and visually coherent region-based stylization than existing blending strategies.

\subsubsection{Quantitative evaluations}
\noindent\textbf{Evaluation metrics.}
To complement the above visual comparisons, we design a set of metrics that directly reflect the objectives of our IGBR algorithm. Since ground-truth multi-region stylized images are not available, we evaluate boundary quality using task-specific no-reference metrics. Boundary Seam Score (BSS) measures local photometric discontinuity across the object-background boundary. $\Delta_{\text{NI}}$ evaluates the consistency between boundary pixels and reliable interior pixels, while $L_{\text{mix}}$ measures inter-object style leakage in feature space. $E_{\nabla}$ captures excessive boundary gradient energy caused by halos or ringing artifacts, and $E_{\text{int}}$ measures whether the deep stylized interior remains close to the preliminary stylized composite. Lower values indicate better performance for all metrics. The complete mathematical definitions and implementation details are provided in the supplementary material.

\begin{table}[H]
\centering
\caption{\small Quantitative comparison of boundary handling}
\label{evaluation}
\setlength{\tabcolsep}{6pt}
\begin{adjustbox}{max width=\linewidth}
\begin{tabular}{lccccc}
\toprule
Method 
& BSS $\downarrow$ 
& $\Delta_{\text{NI}} \downarrow$ 
& $L_{\text{mix}} \downarrow$
& $E_{\nabla} \downarrow$ 
& $E_{\text{int}} \downarrow$ \\
\midrule
Deep Blending 
& \textbf{27.16} 
& 0.23 
& 0.32 
& 1.39 
& 0.17 \\

Poisson Blending 
& 31.42 
& 0.31 
& 0.39 
& 1.82 
& 0.11 \\

Alpha Blending 
& 44.08 
& 0.44 
& 0.10 
& 1.71 
& 0.02 \\

IGBR (Ours) 
& 32.05 
& \textbf{0.06} 
& \textbf{0.03} 
& \textbf{1.02} 
& \textbf{0.01} \\
\bottomrule
\end{tabular}
\end{adjustbox}
\end{table}

Table~\ref{evaluation} quantitatively supports the visual comparisons. We can see that although Deep Blending and Poisson Blending obtain lower BSS due to stronger smoothing, they suffer from higher style leakage, gradient distortion, or weaker interior preservation. More specifically, Deep Blending obtains the lowest BSS, indicating strong seam smoothing, but this comes with higher style leakage ($L_{\text{mix}}$) and weaker interior preservation ($E_{\text{int}}$), suggesting unconstrained smoothing across semantic regions. Poisson Blending also achieves competitive BSS, yet its higher $\Delta_{\text{NI}}$, $L_{\text{mix}}$, and $E_{\nabla}$ indicate color drift and cross-region diffusion under heterogeneous styles. Alpha Blending preserves interior stylization relatively well, but produces the highest BSS and boundary inconsistency, showing that simple alpha transitions are insufficient for incompatible object-background appearances. In contrast, IGBR achieves the best results on $\Delta_{\text{NI}}$, $L_{\text{mix}}$, $E_{\nabla}$, and $E_{\text{int}}$. Although its BSS is not the lowest, it provides a better trade-off between seam suppression, semantic isolation, and stylization preservation. This validates the effectiveness of formulating boundary handling as a region-wise constrained refinement rather than unconstrained blending. 

\noindent Apart to the above analysis, we clarify the difference between our method and generative or harmonization - based baselines. This comparison is presented in the supplementary file.

\subsection{Ablation study}
\begin{figure}[H]
    \centering
    \includegraphics[width=\linewidth]{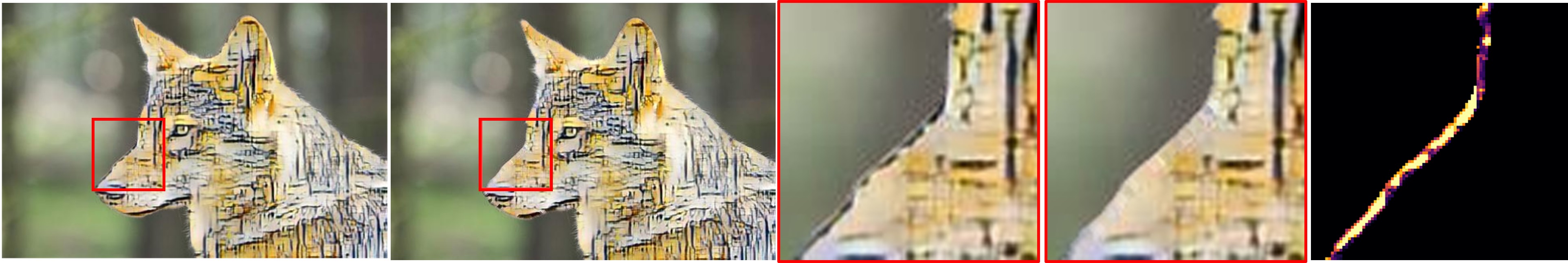}
    \caption{\small From left to right: result of feather-only, result of IGBR, zoom-in of (a), zoom-in of (b), amplified difference between (c) and (d)}
    \label{fig_ablation_study}
\end{figure}

We further visualize the contribution of the repair step in Fig.~\ref{fig_ablation_study}. To isolate its effect, feather-only and full IGBR are compared under the same feathering width, differing only in whether interior-guided repair is applied before feathering. Feather-only smooths the object-background transition but still blends unreliable stylized boundary pixels, which may produce local dark or colored halos. In contrast, full IGBR first replaces boundary-adjacent pixels with reliable interior appearance, yielding a cleaner transition before distance-based feathering. The amplified difference map shows that the repair effect is localized around the object boundary while preserving the stylized interior, confirming its role in stabilizing the feathering zone and preventing boundary contamination.

Table~\ref{tab:ablation_IGBR} analyzes the contribution of each IGBR component. Feather-only greatly reduces BSS and $E_{\nabla}$ compared with Repair-only, confirming that inward feathering is the main factor for suppressing hard boundary seams. However, its higher $\Delta_{\text{NI}}$ and $L_{\text{mix}}$ indicate that directly blending unreliable stylized boundary pixels can still produce local halos or color contamination. Repair-only better preserves interior appearance, yielding lower $\Delta_{\text{NI}}$ and $E_{\text{int}}$, but its high BSS and $E_{\nabla}$ show that repair alone cannot ensure a smooth object-background transition. Full IGBR combines both effects, stabilizing boundary pixels before feathering and achieving the best overall trade-off among seam reduction, semantic isolation, and stylization preservation.

\begin{table}[h!]
\centering
\caption{\small Analysis of ablation study}
\label{tab:ablation_IGBR}

\setlength{\tabcolsep}{5pt}
\renewcommand{\arraystretch}{1.12}

\begin{adjustbox}{max width=\linewidth}
\begin{tabular}{lccccc}
\toprule
\textbf{Variant} 
& \textbf{BSS $\downarrow$} 
& $\boldsymbol{\Delta_{\mathrm{NI}} \downarrow}$
& $\boldsymbol{L_{\mathrm{mix}} \downarrow}$ 
& $\boldsymbol{E_{\nabla} \downarrow}$
& $\boldsymbol{E_{\mathrm{int}} \downarrow}$ \\
\midrule

Full IGBR 
& $\mathbf{34.126 \pm 3.12}$ 
& $0.06$ 
& $\mathbf{0.03}$ 
& $\mathbf{1.02}$ 
& $0.01$ \\

Feather-only 
& $35.041 \pm 3.86$ 
& $0.11$ 
& $0.05$ 
& $1.10$ 
& $0.02$ \\

Repair-only   
& $55.934 \pm 8.21$ 
& $\mathbf{0.05}$ 
& $0.04$ 
& $1.82$ 
& $\mathbf{0.00}$ \\

\bottomrule
\end{tabular}
\end{adjustbox}
\end{table}

\begin{figure}[t]
    \centering
    \includegraphics[width=0.8\linewidth]{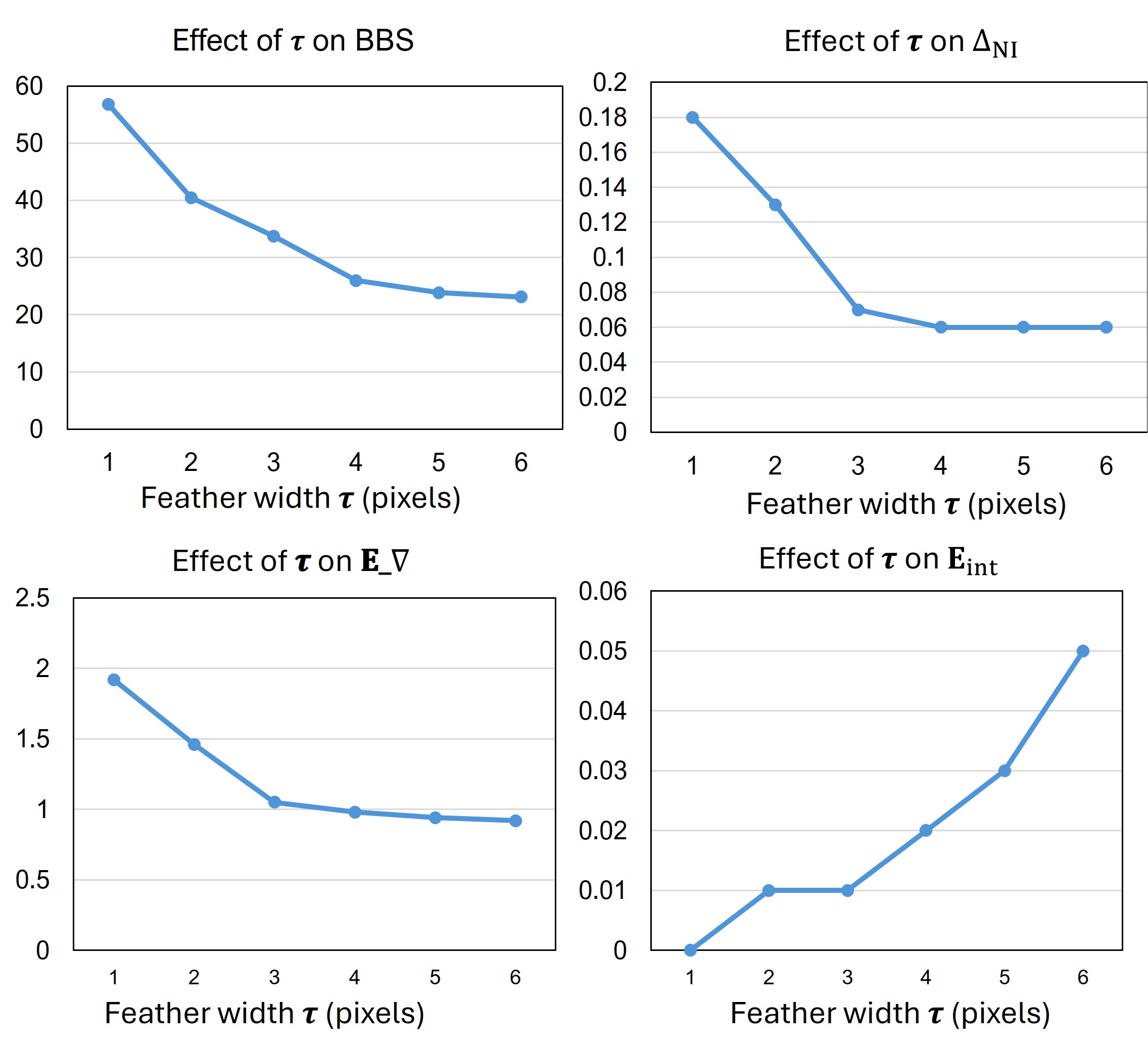}
    \caption{\small Analysis of the sensitivity of the feathering width}
    \label{fig_ablation_metrics}
\end{figure}

To further evaluate the sensitivity of the feathering width, we vary $\tau$ while keeping the border thickness fixed at $3$ pixels. The analysis results are presented in Fig.\ref{fig_ablation_metrics}. We can see that increasing the feathering width $\tau$ consistently improves boundary smoothness. BSS drops sharply from $56.78$ at $\tau=1$ to $26.00$ at $\tau=5$, indicating that a very narrow feathering band is insufficient to suppress hard seams. After $\tau=5$, the improvement becomes marginal, with BSS decreasing only from $23.85$ to $23.12$ when $\tau$ increases from $8$ to $12$. Meanwhile, $E_{\text{int}}$ gradually increases, suggesting that overly large feathering widths may slightly affect interior stylization. Therefore, $\tau=8$ provides a practical trade-off between seam reduction and stylization preservation, while $\tau=12$ gives the lowest boundary-related scores.

\noindent \textbf{Limitation.} The limitation of the proposed framework arises when stylization is applied primarily to the background rather than to foreground objects. Its performance may degrade in background-centric stylization, where the background is stylized while foreground objects are preserved; in such cases, inward feathering may propagate background appearance toward fine object boundaries and reduce semantic coherence. Detailed visual examples and analysis of this limitation are provided in the supplementary material.

\section{Conclusion}
In this work, we address boundary artifacts in multi-region style transfer, where independently stylized regions often produce halos, over-smoothing, and inter-object style leakage after recomposition. We formulated boundary handling as a region-wise constrained refinement problem and proposed IGBR, a lightweight and model-agnostic algorithm that combines interior-consistent repair with inward feathering along the union boundary of stylized regions. The method admits a closed-form solution and can be integrated with existing stylization pipelines. Experiments and evaluations show that IGBR produces cleaner and more stable boundaries than classical blending methods, while ablation studies verify the role of the repair and feathering components. Future work will explore adaptive boundary modeling and semantic weighting for background-centric stylization scenarios.

\bibliographystyle{spmpsci}
\bibliography{Ref}
\end{document}